\documentclass{article}
\usepackage{graphicx}
 
 \usepackage[final]{nips_2017}

\usepackage[utf8]{inputenc} % allow utf-8 input
\usepackage[T1]{fontenc}    % use 8-bit T1 fonts
\usepackage{hyperref}       % hyperlinks
\usepackage{url}            % simple URL typesetting
\usepackage{booktabs}       % professional-quality tables
\usepackage{amsfonts}       % blackboard math symbols
\usepackage{nicefrac}       % compact symbols for 1/2, etc.
\usepackage{microtype}      % microtypography

\title{Handwritten and Machine printed OCR for Geez Numbers Using Artificial Neural Network}

\author{
  Eyob Gebretinsae Beyene \\
  Addis Ababa University, AAiT, ITSC\\
 Addis Ababa, Ethiopia\\
 \\
  \texttt{eyobmbt@gmail.com} \\
}

\begin{document}

\maketitle

\begin{abstract}
 Researches have been done on Ethiopic scripts. However studies excluded the Geez numbers from the studies because of different reasons. This paper presents offline handwritten and machine printed Geez number recognition using feed-forward back propagation artificial neural network.
  On this study, different Geez image characters were collected from google image search and three persons are instructed to write the numbers using pencil. In total we have collected 560 numbers of characters. We have used 460 of the characters  for training and 100 are used for testing. Accordingly we have achieved overall all classification $\sim$$89.88\%$
\end{abstract}

\section{Introduction}

In Ethiopia different Semitic languages are spoken. Semitic languages like Tiginrya, Guraginya and the official working language Amharic are widely used in Ethiopia. Researchers refer to such languages as Ethiopic languages [2] .Language studies [1] upkeep that, the root of all the Ethiopic languages is Ge’ez and currently it survives in church liturgy and literature. Starting form the Aksumite kingdom Geez was the national language of Ethiopia for over centuries [1]. 

On one hand, hundred years back Ethiopian history had been written in Geez for centuries. Though, the language is now working on Ethiopian Orthodox Church only, but enormous amount of Geez manuscripts are existing across the country. Such manuscripts was written using ink and quill. In order to keep the Ethiopian history, it is essential to digitize such manuscripts in to text format.  On the other hand, due to Amharic is an official working language of Ethiopia, researches on Ethiopic languages focuses [2] on recognition of Amharic text. Though, Amharic use almost all the alphabets of Geez but normally it excludes Geez numbers and use Arabic numerals. In our view it is essential to digitize and to research Geez numbers, since we can find the symbols on different types of historical manuscripts. In addition Geez numbers has been commonly used in Ethiopian calendars, Tigiringa and Amharic Bibles and local publications.

Geez numbers represents by 1-9 symbols but there is no zero concept and as such no symbol for zero. However, there are independent symbols for 10, 20, 30, 40, 50, 60, 70, 80, 90, and 100. Figure 1 shows the list of Geez numbers with the corresponding Arabic numeral or English number equivalent.The  Geez character numbers have symbol similarities with the Geez letters. The main difference in most of the symbols are the two dashes/horizontal lines above and below the character of Geez. Figure 2 shows sample selected characters how the Geez numbers have similarity with the Geez letters.
   \graphicspath{ {./image/} }
\begin{figure}[h]
  \centering
     \includegraphics{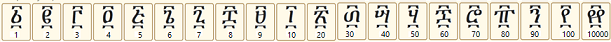}
  \caption{Geez number representations}
\end{figure}

   \graphicspath{ {./image/} }
\begin{figure}[h]
  \centering
 \includegraphics{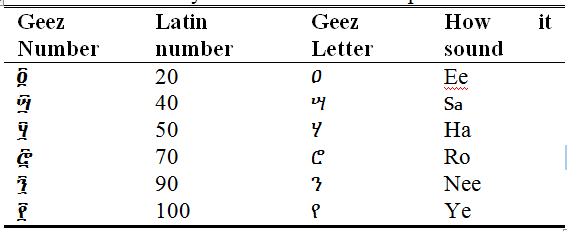}
  \caption{Similarity Geez letters and Alphabets}
\end{figure}

From Figure 2 we comprehend that, researches done on the recognition of Geez/Amharic letter characters will affect the recognition performance of the letters when Geez number are incorporated. This study focused on the recognition of scanned manual and machine printed Geez numbers using artificial neural network.

Humans are gifted and can easily sense and recognize written textual information, however it is still a challenge to detect and recognize textual information by artificial machines [2].Text detection and recognition from scanned images, photos, and different posters is an interesting and challenging research area in computer vision and machine vision [3]. According to [4][5], sufficient of researches have been done on optical character recognition systems on Latin characters. According [1], OCR has been an active research area for more than three decades, however the lack of precision in historical documents recognition is still an issue. Moreover this problem is much greater in script languages of economic developing countries. However, it is fact that researches done on recognition of computer printed documents have greater accuracy in detection and recognition [3][4]. 

However, there are few researches done on OCR for Ethiopic scripts [4][5][6]. Due to insufficient researches on Ethiopic scripts, there is no system or tool for Ethiopic Script recognition. Hence, this research will have its own contribution on building knowledge and method on Ethiopian script recognition. The remaining topic of this paper is organized as follows. We present overview of Neural network and its architecture on  topic two, design and implementation of feed-forward back propagation neural network on topic three, result and discussion on part four and conclusion of the study on part five. 

\section{OCR and Neural Network}
Artificial neural network use number of processing nodes called neurons linked in connected layers [7].  Each node perform some computation and each link have assigned weights and used train the network by adjusting such weights [8]. Researches have been done on both handwritten and scanned machine printed of different language texts. Articles on optical character recognition use different techniques and methods to convert characters to texts such us Neuro-Fuzzy Model and Artificial Neural Network [3][7][8] .Studies on Ethiopic character recognition recommend researchers to apply artificial neural network as a future study direction in order to increase recognition performance [6]. 
From the different neural network architectures feed-forward network architecture is commonly used for character recognition [7] [9]. Such network are used a number of neurons as input layer, hidden layer and output layer. On this study we apply multi layer back-propagation feed-forward neural network architecture
\section{Design and Implementation}
\subsection{Data Collection}
As this study focus on both handwritten and machine printed Geez number recognition, scanned images of Geez number characters are collected from google image repository and from bible. Since limited images are available on google image repository, typed Geez number characters are incorporated to the data set using regular, italic and bold styles.  In addition handwritten Geez character numbers are collected by scanning written numbers which was instructed to three individuals’ to write the numbers using pencil. Sample collected images are shown in Figure 3.
Accordingly 560 amount of data was collected from different sources. Since we select the 20 distinct symbols of Geez numbers, we have prepared 28 sample data set for each character characters. Independent scanned Geez number of characters are stored on a file. We use $82.2 \%$ of the data for training and  $17.8 \%$ for testing. Each file is acquired by the system for further processing. Figure 4 and Figure 5 presents each process before the image is ready for the neural network.
\subsection{Image PreProcessing}
The collected Geez number characters are loaded as independent images. Each image is then ready for the next step which is image pre processing. Character images are converted to gray scale image. The gray images then converted to binary image. Before applying segmentation we inverse the image in order to have one (1) value for the character. 
\subsection{Character Segmentation and Resize}
Since the characters are placed on different area of the image, character segmentation is required. Accordingly we apply a method to crop the character based on the first and last row and column one (1) value of the image. Moreover, for having the same number of pixel on each character we apply a function to resize the image to 45x40. 
   \graphicspath{ {./image/} }
\begin{figure}[h]
  \centering
 \includegraphics{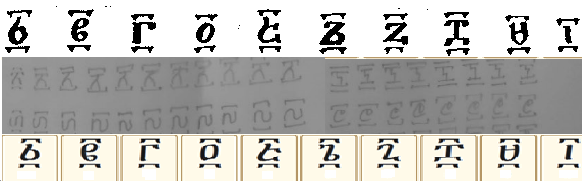}
  \caption{Sample Collected images}
\end{figure}

\begin{figure}[!tbp]
  \centering
  \begin{minipage}[b]{0.3\textwidth}
    \includegraphics[width=\textwidth]{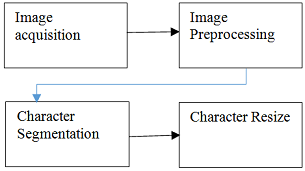}
    \caption{Segmentation Process}
  \end{minipage}
  \hfill
  \begin{minipage}[b]{0.3\textwidth}
    \includegraphics[width=\textwidth]{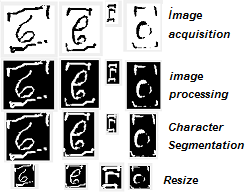}
    \caption{Segmented Characters}
  \end{minipage}
\end{figure}
\graphicspath{ {./image/} }
\begin{figure}[h]
  \centering
 \includegraphics{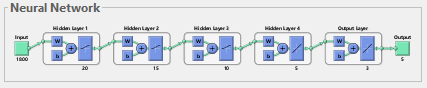}
  \caption{Layers of the neural network }
\end{figure}

\section{Result and Discussion}
\subsection{Implementation}
The feed-forward back-propagation network is implemented using Matlab. The input layer have inputs of the image matrix 45x40 which is 1800. Each image is converted to a single column matrix array and feed to the network. The network have 3 hidden layers with 20, 15 and 10 neurons respectively as shown on Figure 6. The output layer have 20 neurons that represent the 20 Geez number symbols. Output numbers are represented using 5 sequential binary numbers. On this study we apply training function that updates weight and bias values according to conjugate gradient back propagation with Polak-Ribiére updates. The overall classification and recognition performance of the system is achieved  $89.88 $\%. This classification result is restricted by the amount of the collected training and test data sets. However better recognition performance can be achieved when the data set is more than the collected data set on this study. Figure 7 show the overall classification performance of the neural network.
\subsection{Training and Testing}
The network is trained using 460 training data set, which represent 23 independent characters for single Geez number symbol. Accordingly we achieved $98.03 $\% training classification performance. This indicates that, the feed-forward back propagation training classification result is outstanding.  Figure 8 show the training classification result of the network. 

The network is tested using 100 training data set, which represent 5 independent characters for single Geez number symbol. Based on the test data we have, the network was tested and achieved a good result which is $65.3$ \% accuracy. Figure 9 show the test result of the network.
\section{Conclusion}
This paper implement a feed-forward back propagation network with four hidden layer. The paper presented how Geez number character images prepared for the training and test data sets, pre processing on the collected images, training and testing of the neural network.  Accordingly we achieved $89.88$ \% of overall classification and recognition performance. 
\begin{figure}[!tbp]
  \centering
  \begin{minipage}[b]{0.3\textwidth}
    \includegraphics[width=\textwidth]{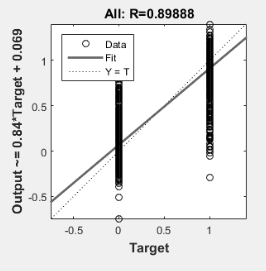}
    \caption{Recognition result}
  \end{minipage}
  \hfill
  \begin{minipage}[b]{0.3\textwidth}
    \includegraphics[width=\textwidth]{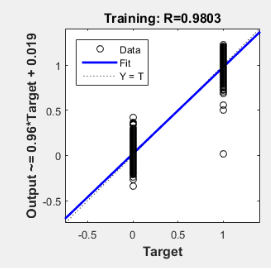}
    \caption{Training Result}
  \end{minipage}
  \hfill
  \begin{minipage}[b]{0.3\textwidth}
    \includegraphics[width=\textwidth]{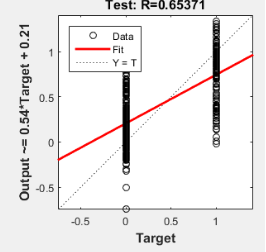}
    \caption{Test result}
  \end{minipage}
\end{figure}
\clearpage
\section*{References}
\small

[1]	W. K. Chen, Linear Networks and Systems. Belmont, CA: Wadsworth Press, 2003.

[2]	Yaregal Assabie, J. B. , “Ethiopic Character Recognition Using Direction Field Tensor”, The 18th International Conference on Pattern Recognition (ICPR'06), IEEE.,2006.

[3]	Sachin Kumar, H. S. B., Annapurna Singh, "A Review on Character Recognition and text Detection: Neural-Fuzzy Approach." Journal of Basic and Applied Engineering Research, 2014, Vol 1, pp. 61-64 5.

[4]	JOHN COWELL, F. H., “Amharic Character Recognition using a Fast Signature Based Algorithm”, Proceedings of the Seventh International Conference on Information Visualization, IEEE,2003.

[5]	Million Meshesha, C. V. J., "Optical Character Recognition of Amharic Documents ", African Journal of Information and Communication Technology, vol. 3, pp. 53-66, 2007.

[6]	Yaregal Assabie, J. B., "Lexicon-based Offline Recognition of Amharic Words in Unconstrained Handwritten Text." IEEE, 2008.

[7]	Chirag I Patel, R. P., Palak Patel., "Handwritten Character Recognition using Neural Network." International Journal of Scientific \& Engineering Research vol 2, pp.1- 6, 2011

[8]	Gunjan Singh, A. P., Sushma Lehri, "Neuro-Fuzzy Model Based Classification of Handwritten Hindi Modifiers " International Journal of Application or Innovation in Engineering \& Management (IJAIEM) vol.3, pp.311-325,2014.

[9]	J.Pradeep, E. S., S.Himavathi. "Diagonal Based Feature Extraction For Handwritten Character Recognition System Using Neural Network." IEEE, pp .364-368, 2011

\end{document}